# Research and Education Towards Smart and Sustainable World

**JUKKA RIEKKI**[ORCID]**1, (Member, IEEE), AND AARNE MÄMMELÄ**[ORCID]**2, (Senior Member, IEEE)**
[1]Center for Ubiquitous Computing, University of Oulu, 90570 Oulu, Finland
[2]VTT Technical Research Centre of Finland Ltd., 90570 Oulu, Finland

Corresponding author: Jukka Riekki (jukka.riekki@oulu.fi)

This work was supported in part by the Academy of Finland through the 6Genesis Flagship project under Grant 318927, and in part by the VTT Technical Research Centre of Finland Ltd. through the Communications in the Smart World (COMMIT) project.

**ABSTRACT** We propose a vision for directing research and education in the field of information and communications technology (ICT). Our Smart and Sustainable World vision targets prosperity for the people and the planet through better awareness and control of both human-made and natural environments. The needs of society, individuals, and industries are fulfilled with intelligent systems that sense their environment, make proactive decisions on actions advancing their goals, and perform the actions on the environment. We emphasize artificial intelligence, feedback loops, human acceptance and control, intelligent use of basic resources, performance parameters, mission-oriented interdisciplinary research, and a holistic systems view complementing the conventional analytical reductive view as a research paradigm, especially for complex problems. To serve a broad audience, we explain these concepts and list the essential literature. We suggest planning research and education by specifying, in a step-wise manner, scenarios, performance criteria, system models, research problems, and education content, resulting in common goals and a coherent project portfolio as well as education curricula. Research and education produce feedback to support evolutionary development and encourage creativity in research. Finally, we propose concrete actions for realizing this approach.

## I. INTRODUCTION

We are convinced that common research paradigms and goals, including shared visions and research problems, promote teamwork and result in coherent project portfolios, as suggested in [1], [2], and hence accelerate research and development. Our selection for the *core goal* is to provide prosperity for the people and the planet through better awareness and control of the environment, both natural and human-made. We base the goal on the United Nations 2030 sustainable development agenda [3] and its 17 Sustainable Development Goals (SDGs). The goal is in line with the European Union's strategic agenda 2019-2024 and its main priority, *"building a climate-neutral, green, fair and social Europe"* [4], leading to the European Green Deal [5].

Brundtland Commission (1987) provided an excellent definition of sustainability: *"Sustainable development is development that meets the needs of the present without compromising the ability of future generations to meet their own needs"* [6]. A pioneer in promoting sustainable development has been the Club of Rome, founded in 1968, as a discussion forum by Aurelio Peccei and Alexander King, leading to the report *Limits of Growth* in 1972, later updated [2]. The Earth overshoot day is alarming evidence of the problems in global development [7]. We agree that exponential economic growth cannot continue because of finite resources. Instead, the focus has to move from quantitative to qualitative improvement.

Starting from the core goal, we formulate a vision for guiding research and education in information and communications technologies (ICT). We select the *Smart World* vision [8], [9] as the basis due to its general nature. This vision is originally based on Mark Weiser's idea of ubiquitous or pervasive computing [10], which was preceded by the ideas of distributed systems and mobile computing [11]. The term Smart World has been used in the literature for decades, but it was first linked to Weiser's vision in [8], [12]. Moreover, our formulation is inspired by the *Smarter Planet* vision [13]: the world will be *instrumented* using sensors and actuators, *interconnected* using communications, and *intelligent* using traditional optimization, conventional artificial intelligence (AI), and computational intelligence (CI). We often use the term AI to correspond to conventional AI and CI.

We define the Smart and Sustainable World vision as follows: *Prosperity for the people and the planet is achieved with intelligent systems that sense their environment, make proactive decisions on actions advancing their goals, and perform the actions on the environment. Sustainable development is emphasized in decision making, and system performance is optimized to save basic resources. Humans observe the autonomous operation through user interfaces and, when needed, revise the operation or control the systems manually.*

The term Smart and Sustainable World has been used earlier, for example, in [14], but has not been defined as above. Our vision is empowered by advances in technology that introduce ICT devices and intelligent systems everywhere. As an example, miniaturization, wireless technologies, and energy harvesting enable self-powered sensors and user interfaces to be embedded in everyday objects. This technology and its increasing intelligence need to fulfill the needs of society, individuals, and industries but also optimize resource usage as intelligent use of the limited basic resources will be crucial to realize sustainability. This apparent requirement is recognized in [14] and [15], for example. The role of intelligent systems (i.e., artificial intelligence) in achieving SDGs is recognized in the EU White Paper on AI as well [16]. Technology acceptance and human control are essential to realizing the vision. Moreover, intelligent systems must act according to our society's cultural values, norms, and laws. Intelligent systems and human beings are discussed in Section II. Here, our discussion on building sustainability and guiding AI research and development is based mainly on reports from the EU as it has been recently quite active in these fields.

The vision leads to truly complex systems of systems [17]: a large number of interconnected devices provide a distributed platform for numerous co-existing intelligent systems that share the platform's limited resources. We argue that managing the complexity calls for studying system-level research problems - and this requires complementing the conventional atomistic *reductive view* for conducting research with a holistic *systems view* [18]–[20].

In reductive thinking, also called analytical thinking, the idea is to start from conceptual analysis, reduce research problems to simpler problems, then perform experiments, generalize the results to a theory by induction, abduction, or formation of a hypothesis, and finally derive results to the original problem from the theory by deduction. A system is assumed to be the sum of its parts. However, this is valid only in linear systems that follow the superposition principle [21]. Other systems are nonlinear. In complex systems, nonlinear relationships between system parts cause a phenomenon called emergence or even chaos [18]. Nonlinear systems should be avoided, but when this is not possible, systems thinking [2] and systems engineering [22] are called for. As systems thinking is more general than reductive thinking, it can cover more complex phenomena. Deduction can be replaced with some more general nonlinear inference when deriving the result for complex systems. Moreover, simulations and experiments with system prototypes provide insight into system-level properties.

The growing need for systems thinking to solve complex problems is in line with the general progress of science from simple to complex. For example, the short history of mathematics of the last century included in [23] shows this development in mathematics. Systems thinking is needed in the mission-oriented research that targets common goals based on the United Nations sustainable development agenda and the European Union's strategic agenda [3], [4]. Further, systems thinking calls for inter- and transdisciplinary research and thus helps to avoid the fragmentation of science. General arguments for systems thinking and inter- and transdisciplinarity include answering complex questions, addressing broad issues, solving problems beyond the scope of any one discipline, exploring disciplinary and professional relations, and transferring knowledge between disciplines [24], [25]. The need for systems view has been recognized widely, for example in [26]–[29], [30]–[33], [34]–[37], [38]–[40].

As a summary, we suggest the research paradigm of combined reductive and systems thinking for studying complex systems. One of 12 places to intervene in complex systems are paradigms [2] and the other two emphasized by us are goals and self-organization. Goals include visions and research problems, and self-organization includes independent research based on the research paradigm. We present detailed arguments for the suggested research paradigm in Section III.

We identify scenarios, system requirements, performance parameters, basic resources, system ideas, and system models as the key concepts to consider when specifying intelligent systems. The disciplines related to the Smart and Sustainable World vision have similar *system ideas* such as *optimization*, *decision-making*, *open-loop* and *closed-loop (feedback) control*, *hierarchy*, and *degree of centralization* [41], [42]. Hierarchical systems are modular. Specifically, feedback loops formed of sensors, decision blocks, actuators, and controlled processes are crucial for intelligent behavior. The subsidiarity principle guides in organizing systems into hierarchical and relatively autonomous subsystems. These topics are discussed in Section IV.

We suggest a systematic approach for planning research and education based on the Smart and Sustainable

World vision. This approach supports setting system-level research problems and applying systems thinking to solve them. We sketch steps starting from the vision and proceeding through scenarios and requirements to system specification, and finally to research problems and education content. These steps result in a preliminary system model, a set of research problems, and content for the education curricula. The research problems form a basis for a coherent project portfolio. The curricula, in turn, lay a basis for educating researchers with the required general knowledge to contribute towards the vision. We present the steps in Section V. In Section VI, we suggest actions to start aligning research and education with the vision. Section VII presents the conclusions.

To serve a broad audience, we explain in each section the essential concepts and list key literature. As the literature on these topics is quite fragmented, this survey can help interested readers locate relevant research, avoid work already done, and apply the latest knowledge in their research.

## II. INTELLIGENT SYSTEMS AND HUMAN BEINGS
### A. INTELLIGENCE
Increasing the intelligence of human-made systems is at the core of the Smart and Sustainable World vision. In standard language, intelligence implies human-level self-consciousness that indicates *"having or showing awareness of one's own existence, actions, etc."* [43]. Human beings have the unique property of reflective consciousness [44]. They are aware of their sensations and their environment in a broad sense, know history, and plan and anticipate the effects of their actions.

As our focus is on intelligent human-made systems, we follow the definition of intelligence in general engineering terms [45], [46]: *"the ability of a system to act appropriately in an uncertain environment, where appropriate action is that which increases the probability of success, and success is the achievement of behavioral subgoals that support the system's ultimate goal. Both the criteria of success and the system's ultimate goal are defined external to the intelligent system."* The goal of an intelligent system may correspond to some human need [47]. Uncertainty means lack of precise knowledge. Intelligence can also be defined as *"a ratio, or quotient, of the ability of a system to control its environment versus the tendency of the system to be controlled by the environment"* [48]. This is a useful definition for our vision, as it emphasizes controlling.

Performing appropriate actions in an uncertain environment is central for intelligence and leads to the need of control systems and feedback [46], [48], [49]. Feedback is a key concept in several well-known models such as the *sense, act, and plan* paradigm, the *cognitive* cycle, the *observe, orient, decide, and act* (OODA) loop, and the *monitor, analyze, plan, execute, and knowledge* (MAPE-K) loop [41]. Control systems have been expanded to achieve intelligence and manage uncertainty by applying traditional optimization, conventional AI, and CI to learn from and make decisions based on the state of the environment observed via sensors and by changing the state of the environment based on the decisions via actuators.

Traditional optimization methods include exhaustive search, local search, linear programming, dynamic programming, divide and conquer, scalarization, cybernetics (based on feedback), and game theory [50], [51]. However, the global optimum (i.e., the goal) is not usually easy to find because of many variables, dynamic environment, nonlinearity of the problem, local optima, and limited observability and controllability of the environment or process to be controlled [52].

Artificial intelligence is sometimes called machine intelligence [53]–[56], [49], [57]. Conventional AI is based on symbol manipulation, hard computing methods using binary deductive logic, and statistical methods. CI is using pattern recognition and soft computing methods. AI includes expert systems, case-based learning, statistical learning, and multiagent systems. CI includes statistical methods, neural networks, fuzzy systems, and evolutionary computing (e.g., genetic algorithms and swarm intelligence) [58]. Some authors exclude statistical pattern recognition methods from CI. Conventional AI such as expert systems and CI such as neural networks can be combined into hybrid intelligent systems [59], [60] or a fusion of hard and soft computing [61].

Pattern recognition is a general principle that integrates various other principles. Information can be defined as patterns or structures, and information transmission is, in principle, transmission of patterns [62], [63]. The state of a system as a set of system properties [64] can also be seen as a pattern. The human brain is a large neural network based on a complement to deductive logic, namely pattern recognition that is not as sensitive to errors [48], [65]. Evolution, self-organization, and the use of genetic algorithms are basically formation of patterns. Pattern recognition has produced such fields as data mining and big data. Many principles can be seen as special cases of pattern recognition, for example, data detection and channel estimation in communications, feature detectors in radio signal classification and interception, and formation of situation awareness. When pattern recognition (analysis) and formation (synthesis) are combined with feedback, they form a decision block of an intelligent (in fact, learning) system [66]. A learning system changes its behavior based on earlier experience [41]. More generally, a learning system not only remembers but also can generalize beyond its own experience to new situations [67].

The term intelligence is often replaced in engineering with the more technical term *rationality* since often intelligence refers to analytical intelligence as in intelligence quotient (IQ) tests and not directly to the ability to attain goals [68]. Furthermore, rationality includes the idea of resource efficiency. AI may be defined as a theory of rational agents where an agent is *"any device that perceives its environment and takes actions that maximize its chance of success at some goal"* [49]. A recent definition of AI resembles closely the definition of intelligence above: *"Artificial intelligence (AI) refers to systems that display intelligent behaviour by analysing their environment and taking actions – with some degree of autonomy – to achieve specific goals"* [69]. Edge AI, that is, distributing AI to the edge of the network, near the data sources, humans, and controlled processes, widens the application area of AI by shortening latency or delay, enhancing privacy, and improving performance [70], [71].

### B. HUMAN ACCEPTANCE AND CONTROL

When intelligent systems become more common, humans use them to an increasing amount to consume the services they need and contribute to society. In addition to intelligent systems operating in the physical environment (vehicles, factories, service robots, and household appliances, for example), many intelligent systems operate as embedded systems and control the information infrastructure and the services. These systems decide and learn where to locate computations and how to share the communication resources to maintain a smooth user experience for social media, for example. In addition, these systems decide the content that services show for humans and also use services on behalf of humans.

Introducing such pervasive decision power for intelligent systems requires considering how humans accept these systems and stay in control. Broadly, these systems can be accepted when they are useful, easy to use, and ethical. Useful services fulfill the needs of humans [72], and services are easy to use when they have high usability [73]. Regarding ethics, the systems have to behave according to the cultural values, norms, and laws of our society, including safety, security, cooperation, trust, justice, integrity, respect, and privacy.

Perceived usefulness and perceived ease of use are considered central factors in technology acceptance, but additional factors have been listed as well [74], [75]. Of these, we emphasize trust due to the significant role that intelligent systems have in the Smart and Sustainable World. Transparency supports building trust by presenting the grounds of the decisions and the use of data to humans. Explainable or interpretable AI, that is, AI capable of justifying its decisions [76], tackles these challenges. The EU's Ethics Guidelines for Trustworthy AI [28] lists *explicability* as one of the four ethical principles for AI systems. The other three are: *respect for human autonomy*, *prevention of harm*, and *fairness*. EU's White Paper on Artificial Intelligence also emphasizes trust, values, and fundamental rights such as human dignity and privacy protection [16]. As data are central for intelligent systems, handling data needs to be considered with care as well. This issue is discussed in more detail in the European strategy for data [77] but is outside the scope of this article.

We highlight the respect for human autonomy, that is: *"Humans interacting with AI systems must be able to keep full and effective self-determination over themselves"* [28]. This principle is related to the principle of *"meaningful human control,"* stating *"humans not computers and their algorithms should ultimately remain in control"* [78]. Such human authority is essential; humans must have power over human-made intelligent systems. Systems providing meaningful human control are responsible to human moral reasons. Moreover, their operation is traceable to one or more relevant persons with proper moral understanding. These are called the tracing and tracking conditions [78]. Tracking ensures human authority.

The possibility to revise the operation of an intelligent system or control a system manually is important because an autonomous system can generate chaotic behavior if some parts fail and not enough redundancy is used, sensor data are contradictory, or the system does not have a clearly defined goal. Human intervention might also be needed when a system is behaving as designed, but the design or behavior learned by the system has flaws. Even without flaws, automation and autonomy may sometimes create dangerous or surprising situations. Furthermore, variations in the available resources might require manual control, for example, when an autonomous vehicle loses connectivity to the infrastructure.

Acceptance and human control set requirements for the system, specifying, for example, sufficient cybersecurity to prevent unauthorized access and deliberative malfunctions, consent management to determine the operations humans allow for intelligent systems, and user interfaces to monitor and control intelligent systems. We discuss requirements in more detail in Section IV-A. This type of requirements in automatic decision-making are already recognized in the General Data Protection Regulation (GDPR) and also considered to some extent in some EU Member States' laws [79] though the emphasis is generally on services using data instead of intelligent systems performing actions in the environment.

### C. HUMAN AWARENESS

The vision implies greater awareness for humans. They gain situation awareness based on the information delivered by the intelligent systems and command the systems to advance the goals they have set for themselves. When sensors also measure humans, the humans get insight into their condition. Understanding how their decisions affect their own and their environment's well-being can lead to adjusting the goals. In this way, intelligent systems can persuade [80] humans to take sustainable actions.

This awareness can, in the future, be expanded in the form of telepresence where all the senses (for example, sight, hearing, and sense of touch) come into use. The delays are minimized so that the interactions with the environment look almost instantaneous, and humans feel physically present at the remote site. Combining telepresence and autonomous operation produces systems that make decisions autonomously and control the environment based on these decisions but are observed by remote humans that can also revise system operation and give high-level goals for the systems.

## III. REDUCTIVE AND SYSTEMS APPROACHES FOR CONDUCTING RESEARCH

We argue that managing complexity requires complementing the conventional atomistic *reductive view* for conducting research with a holistic *systems view* [18]–[20], [24]. This combination helps to avoid the tendency towards fragmentation as well, instead generating coherent scientific knowledge using unified terminology over disciplines.

### A. REDUCTIVE THINKING

Analysis and synthesis are common methods in science and engineering. Deduction corresponds to analysis. Induction, abduction, and formation of hypotheses correspond to synthesis. Abduction means inference to the best explanation [81]. In practice, abduction is implemented by strong inference using many competing hypotheses [82]. Reductive thinking, also called analytical thinking, is the basis for the success of the Western culture [19], [48]. Reductive thinking starts from conceptual analysis. Next, a research problem is reduced to subproblems, and the subproblems are studied by performing experiments in the real world. Finally, results to the subproblems are generalized to the theory world by induction, and a result to the original problem is derived by deduction (from the theory world to the real world), see Figure 1. The figure was inspired by [81], [83], [84]. In [84], the figure is upside down in a simplified form, whereas our figure is hierarchical so that the abstract concepts are at the top and concrete things are at the bottom as in [81]. Figure 1 resembles Lewin's learning cycle used in education [85] and includes the idea of feedback.

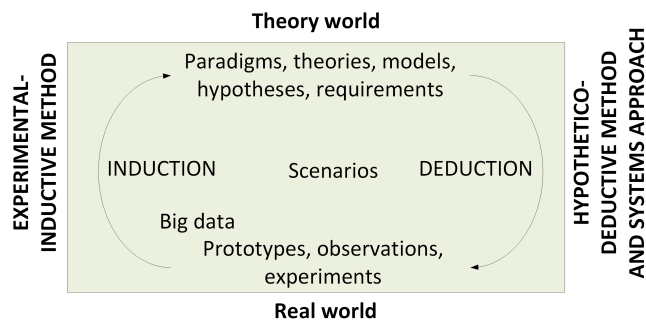

**FIGURE 1.** Structure of science in reductive thinking consisting of the real world in the bottom, the theory world on top, and scenarios in the middle. The structure describes the inductive and deductive parts of both research and learning. The problem is reduced into simpler problems before the inductive and deductive parts. The diagram includes the idea of feedback from the real world to the theory world. The deductive part corresponds to systems thinking but only in mathematically tractable problems.

In science, a hypothesis, theory, or model is verified with the hypothetico-deductive method by deriving results deductively and comparing them with real measurements [86]. In engineering design, we do not initially have any object to be verified, only requirements that do not exist in sciences [22], [87]–[89], [90], [91]. When a complete system is being studied, a system model is built based on the requirements. A system model is an intentionally simplified description of regularities in a system, usually mathematical [86]. The system model is verified by comparing the requirements with results generated through analysis, simulations, or experiments with system prototypes.

In more detail, the research problem corresponds to system requirements, and conceptual analysis produces system specifications, a system model, and finally, a system prototype. This is the hypothesis in various phases. The problem is reduced to subproblems that are studied with subsystems. Generalizing the results refine the hypothesis. Deduction produces predictions about the subsystems. When the whole system can be constructed based on the hypothesis, a full prototype can be first verified with respect to the requirements and then validated in a real environment, i.e., verified whether the system fulfills the needs [90].

In reductive thinking, a whole is assumed to be the sum of its parts, which is valid in linear systems. In complex systems, this assumption cannot always be made because of possible nonlinear relationships between the parts, leading to nonlinear system models and emergent phenomena or even chaos. An old but useful classification of general linear and nonlinear system models is included in [92]. Emergence implies the occurrence of properties at a higher hierarchy level of the system that are not predictable from properties at lower levels. Thus, complex systems' properties cannot be managed with reductive thinking, and systems thinking is hence called for.

### B. SYSTEMS THINKING

In systems thinking, reductive thinking is accompanied by intuition, interdisciplinary discussions, structural analogies, and existing system models known to work. These tools facilitate using transferable knowledge useful in new situations and different disciplines, thus improving our creativity [93]. Reductive thinking is a special case of systems thinking since the latter is also sometimes using analysis.

Intuition is largely based on experience [94]. Interdisciplinary discussions provide knowledge from other disciplines and structural analogies information on possible solutions [2], [93]. The system ideas presented in Section IV-B are applied in the proven system models, for example, to organize a system into loosely coupled relatively autonomous subsystems [47], [95], [96]. These models help to decrease the number of relationships between the parts of the system with subsidiarity (also discussed in Section IV-B).

When deriving the result, deduction can be replaced with some more general nonlinear inference which may be called systems thinking [20]. Moreover, simulations and experiments with system prototypes provide insight into higher hierarchy level properties. They replace the mathematical analysis in cases where the analysis is not possible. Deduction corresponds to systems thinking only in mathematically tractable problems.

In summary, in the combined reductive and systems thinking, bottom-up reductive experimental-inductive research is followed by top-down hypothetico-deductive and systems research. Systems thinking is a form of generalized inference that is needed to replace deduction in mathematically intractable problems. Such problems can be studied by using approved system models and their simulations and experiments.

**TABLE 1.** Comparison of reductive and systems approaches.

| Reductive approach (analytical approach) | Systems approach (holistic approach) |
|---|---|
| Research proceeds bottom up, from simple to complex. | Research proceeds top down. |
| Emphasis is on details at the expense of the whole. | Emphasis is on the whole at the expense of the details. |
| The system is either assumed to be linear or divided into subsystems that are assumed to be essentially uncoupled (Section IV-B3). The system is assumed to be isolated from the environment (Section IV-A2). | Nonlinear weak or strong coupling between the subsystems is considered (Section IV-B3). Hence, emergent phenomena are included in the system model (Section III-A). The whole is seen as part of the environment, and the system is assumed to be open (Section IV-A2). |
| Local optimization, in general, does not lead to global optimization. The focus is often on one objective only. | Analysis is difficult if not possible. The results are usually rough and descriptive. System archetypes and their simulations and experiments must be used to complement deduction (Section VI). |
| Good basis for mathematical analysis. The results are general but valid only within the scope of the assumptions and usually for some simple idealized cases. | Global optimization can improve system efficiency. Mutually conflicting objectives call for trade-offs (Section IV-B2). |
| Traditional scientific method. Must be initially used in education and research. Reductive science tends to be fragmented and use differing terminology. Overlapping efforts cannot be avoided. | Must be used in mission-oriented research for complex systems and to achieve sustainability (Section III-C). Interdisciplinary and transdisciplinary research is emphasized for broad issues (Section III-B). Knowledge can be transferred between disciplines. |
| | Complements the reductive approach. Must be used in education and research after the reductive approach. Fragmentation of research is reduced. |

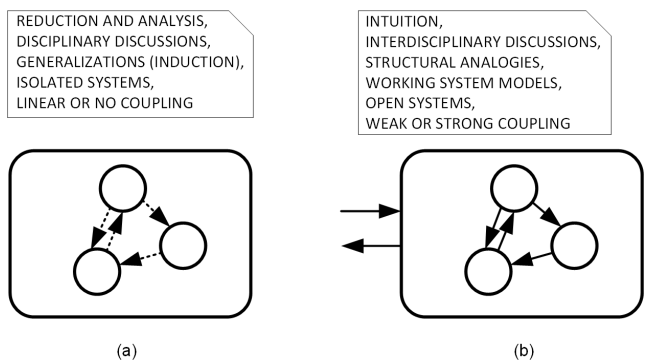

**FIGURE 2.** Different research approaches. (a) Reductive approach emphasizes details, the three subsystems in the figure. Dotted arrows refer to linear or no coupling. b) Systems approach emphasizes the whole, the box enclosing the subsystems. Continuous arrows refer to weak or strong coupling. The arrows on the left side indicate an open system exchanging matter, energy, and information with its environment.

The reductive and systems approaches are summarized in Fig. 2 and Table 1, showing their benefits and limitations. The table provides pointers for the sections describing the terms. When using these principles, one must always accept critical feedback from colleagues to obtain high quality results. A good example of systems thinking is the periodic table of chemical elements. The table integrated the whole chemistry, but its invention was possible only after the elements had been studied reductively. This example shows how one must start with reductive thinking and complement the research with systems thinking.

As intelligent systems are often complex, they call for systems view and studying system-level research problems.

This approach is needed to meet the strict performance requirements in the presence of resource constraints (Section IV-A), specifically when the systems form large-scale systems of systems [17], [97]. Sustainability sets further requirements for resource usage.

Good books on systems thinking include those by Ramage [85] about systems thinkers, Boulding [63] about the hierarchical and evolutionary view of the world as a total system, Meadows about various practical applications of systems thinking [2], Kossiakoff about systems engineering [22], Checkland [20] and Richardson [98] about the history of systems thinking since ancient times, and Bossel [47] about modeling of complex systems. Avizienis *et al.* present an excellent conceptual analysis on dependability and security that have become essential elements in system design, in addition to conventional functionality, performance, and cost [99]. We describe systems view in more detail in [42].

The systems view does not replace the reductive view, but these two complementary views both have their roles in research. Both specialists with their reductive approach and generalists with systems thinking are needed to succeed and minimize risks with the uncertainties of the future [63], [93], [100]. Specialists carry out research using the reductive approach, each in their disciplines. Otherwise, original research is quite difficult to conduct since the amount of knowledge is too large to master in sufficient detail [100]. On the other hand, more experienced researchers using the systems approach can focus the efforts by setting common goals. They can participate in other research activities as well, for example, in defining the research problems and setting the hypotheses. Forecasting is important to have time to adapt to the expected changes. Generalists can be better than specialists in forecasting, even in the specialists' own disciplines, as they have general knowledge and a broad understanding of history to identify relevant future problems [93].

Systems thinking cannot be based only on the reductive *multidisciplinary view* that is an additive view without notable interaction between disciplines [24], [101], [102]. This approach produces new knowledge in separate disciplines and stays within their boundaries. An example is an encyclopedia: experts from different disciplines write articles independently with some guidance from the editor-in-chief. *Interdisciplinary view* is a more advanced, interactive view that analyzes, synthesizes, and harmonizes links between disciplines into a coordinated and coherent whole, producing, for example, a unified report or book on a given research problem [102]. The third and most advanced view is the *transdisciplinary view*: disciplines are integrated into general knowledge in a societal context, and their traditional boundaries are transcended. The final goal is the unity of science. In natural sciences and engineering, the interdisciplinary and transdisciplinary views are applied in systems thinking, i.e., in general system theory and systems engineering.

### C. SYSTEMS THINKING IN MISSION-ORIENTED RESEARCH AND EDUCATION

The United Nations sustainable development agenda and the European Union's strategic agenda lead to the *mission-oriented approach* which also calls for the systems approach to determine the vision and the common goals. The systems approach is also needed to solve the resulting complex problems where the solution can be found only using results of many disciplines [24], [103]. Moreover, the impact of research can be found using the systems approach rather than the reductive approach. *Research platforms* are gaining popularity as a mission-oriented approach to organize research. Research platforms are research programs for interdisciplinary research projects [24], [104] that address large societal and global problems (e.g., related to SDGs) which can only be studied from an interdisciplinary perspective.

The systems approach is needed in education as well (Section V-E). Learning is essentially a bottom-up inductive process, but top-down deduction can strengthen learning [83]. *Learning by doing* corresponds to the reductive approach if the problem is initially reduced to simpler problems following by induction using the bottom up approach. *Integrative learning* with its final top down analysis matches with the systems approach. Proper education curriculum can act as an external force guiding researchers towards systems thinking and to reduce fragmentation.

## IV. DESIGNING INTELLIGENT SYSTEMS
### A. SCENARIOS, SYSTEM REQUIREMENTS AND RESOURCES
#### 1) SCENARIOS AND SYSTEM REQUIREMENTS

*Scenarios* and *use cases* are effective means to discover user needs and define *functional* and *nonfunctional requirements* [105]–[107]. Scenarios have been used as a representation of system requirements to improve communication between developers and users. Scenarios offer a down-to-earth middle-level abstraction between models and reality [106], see Figure 1. We consider use cases to form a special case of scenarios and focus on the type of scenarios that describe the interaction of a system with its environment and users, as well as interaction among its parts.

A *requirement* is a general statement about user needs. Functional requirements define a system or its components, the functions that the system or the components must perform. In contrast, nonfunctional requirements are *performance requirements* that are related to the use of basic resources. The requirements need to concern human acceptance and control as well, for example, to support building trust and providing meaningful human control, as discussed in Section II-B.

We emphasize performance requirements as we expect *intelligent use of basic resources* to be a central condition for realizing the Smart and Sustainable World vision since all the resources are scarce [15], [20], [41]. The world is essentially a closed system regarding materials, and thus their usage should be minimized, and they should be recycled for sustainability [108]. Resource regulation is needed as common and free resources are otherwise used wastefully; this is known as the *tragedy of the commons* [2]. For example, climate change is caused by the wasteful use of the atmosphere.

#### 2) PERFORMANCE AND RESOURCES

The fundamental properties that characterize computing and communication systems are functionality, stability, scalability, performance, dependability, security, and cost [99], [109]. Fulfilling the requirements set for these properties is always challenged by constraints on resource usage. *Performance* and *constraints* are system level challenges, as can be seen already in the definition of intelligence [45]. Thus, they are highly relevant for the vision. Intelligent systems must meet the performance requirements as they will operate in our everyday environment, in close contact with humans, and they will control various processes crucial for the industry and the society. On the other hand, sustainable development places constraints on using basic resources and can challenge the performance requirements.

The *six basic resources* are: *materials*, *energy*, *information* (data and control), *time* (delay), *frequency* (bandwidth), and *space* (size) [41], [63], [110]. The first three basic resources are the most important and well known in general system theory where an *open system* exchanges them with its environment through its boundaries [18], [63], [64], [111], [20]. An *isolated system* does not exchange any matter, energy, or information with its environment [112]. The last three basic resources are well known, for example, in communication theory [113].

Performance often varies in time, frequency, and space, and we are interested in the variations. *Efficiency* is a performance parameter usually measured using *the ratio of benefits and expenditures* [64]. The benefits may be, for example, data bits or operations, and expenditures are usually the consumption of basic resources. The general additional performance parameters are *dependability* and *security* [99], [114]–[116], see Figure 3. Means of achieving dependability include error tolerance, fault tolerance, and robustness. *We identify efficiency, availability, reliability, safety, and security as the essential performance parameters of future systems.* These parameters are emphasized when the technology serves humans, as discussed in Section II.

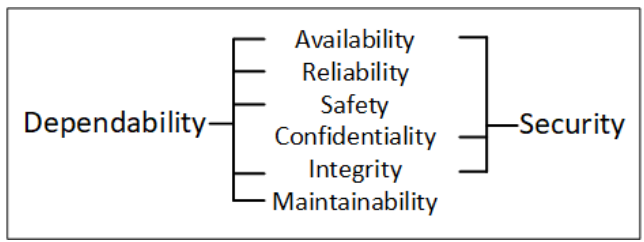

The *fundamental limits of nature* are specific constraints [100], [117], [118]. The continuing development of technology and the ever more stringent requirements of new application areas (requiring, e.g., millisecond-level latency and ultra-low energy consumption) approach these fundamental limits. Thus, understanding system requirements and optimizing resource usage is becoming even more important. Otherwise, the planned services will become expensive or even infeasible to implement. As an example, Moore's exponential prediction of the development of silicon electronics has had problems already since 2004 and will not continue after 2021 [119], [120]. Quantum computing provides an alternative, but it presently needs either cryogenic temperatures or extremely high pressures [121]. Hence, quantum computing cannot, at least in the near future, meet the energy efficiency requirements of many applications.

### B. TRANSDISCIPLINARY SYSTEM IDEAS
#### 1) CLOSED-LOOP FEEDBACK
The closed-loop feedback concept is the basis of most intelligent systems including intelligent agents in AI systems [2], [48], [98], [124], [49], see Figure 4.

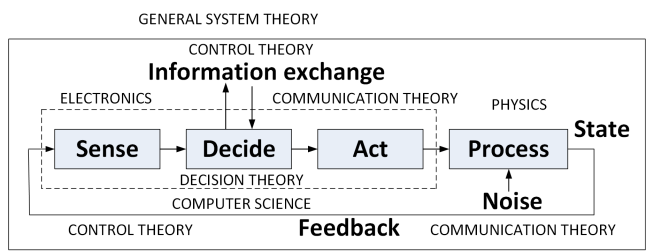

**FIGURE 4.** Basic system model including feedback and some of the related disciplines. The process block is also called the plant or environment.

Feedback includes four blocks, namely the *plant*, *process*, or *environment* to be controlled, *sensing*, *decision making*, and *acting*. Feedback may be positive (reinforcing) or negative (balancing) [2], [98], [124]. Positive feedback may result in exponential growth; it is therefore inherently unstable and needs an outer compensating negative feedback with a definite goal. Systems that do not include the sense block are called open-loop control systems [52]. Systems having no act block are called adaptation or monitoring systems [41], [212]. The adaptation systems do not control the environment using feedback. Many computational intelligence systems are adaptation systems. Data science is also usually based on adaptation systems [53], [57], [122], [123], [216] and may be considered as a subset of artificial intelligence. Although the environment is not controlled, data science may apply feedback as, for example, in reinforcement learning [57]. Such feedback is an essential component of adaptive signal processing, see [212].

Data science can contribute to realizing feedback for intelligent systems as it encompasses the principles that support and guide the extraction of information and knowledge from data [122], [123]. It consists of various, somewhat overlapping concepts and includes machine learning, pattern recognition, data mining, knowledge discovery, and big data. The origins of big data are in pattern recognition, which has a long history in statistics [41]. The first articles in the IEEE literature were published in 1954. Knowledge discovery focuses on methodologies to identify novel and useful patterns from large data sets [122], [123]. Data mining means analyzing data and identifying patterns as the first step in knowledge discovery, but for some authors, data mining and knowledge discovery are synonymous. They started in about 1990. Mashey (1998) suggested the term big data [41], which is concerned with large-volume, complex, ever growing data sets coming from various sources. Big data mean data sets that are too large for traditional data-processing systems and therefore require new technologies [122]. The Internet of Things (IoT) has since 1999 made the available data especially large.

The decision block has the optimization task and it can be *hierarchical* [125]–[127] and *distributed* [128], [129]. In control engineering, the decision block is called the *controller* [52]. In social sciences, the sense, decide, and act blocks correspond to an *actor* [130]. In computer science, they correspond to an *agent* [49].

The goal and the performance criteria are given externally to the decision block. The performance requirements are described numerically using performance criteria. The *goal* is the desired state of the process to be controlled, for example, performance or location. The process is transferred from the present state to the desired state, usually iteratively.

We expect the concepts introduced in this section to be crucial when intelligent resource-efficient systems are developed. *This challenge deserves attention – we need to master feedback to develop the intelligent systems that together realize the Smart and Sustainable World vision.*

#### 2) INTELLIGENT DECISIONS
For intelligent behavior, analyzing, planning, and optimized decisions are needed. The decision block usually includes a more general *planning phase* [46]. The decision block imagines alternative futures using historical information and selects the best one [46], [108]. Thus, an intelligent generalized decision block combines analysis (deductive logic or *pattern recognition*) and synthesis (*pattern formation*, decisions), finally producing control data for the act block [66]. The decision block also includes a model of the process containing the needed *memory* for historical information [46].

The decision block should ideally be not only *reactive* but also *proactive* and in a social environment *cooperative* or *competitive*. According to [46], [131], proactive architectures were first studied by Firby in 1989. Decision making can be challenging when several local optima exist instead of a single global optimum. In *multiobjective optimization*, there may be many conflicting and incommensurate objectives or criteria [51]. Even in the case of commensurate objectives, the global optimum is not unique but a set of optima called *Pareto optima* [51]. Thus, after optimization, a decision must be made with a subjective preference that is usually equity. The problem caused by incommensurability must be solved by an evolutionary process using survivability, in practice, with the law of supply and demand in a free market. However, monopolies might need to be solved separately. We discuss the decision block in more detail in [42].

### 3) HIERARCHICAL SYSTEMS

In a *hierarchical system*, sensing results proceed upwards, and actions proceed downwards. The feedback loops at the upper hierarchy levels are much slower than those at the lower levels [126]. The actions can be understood to form a goal hierarchy so that the upper level actions set goals to the nearest lower levels in the hierarchy. Each level has the priority to set goals to the next lower level to avoid deadlock situations [125].

A system may be *centralized* without any subsystem autonomy or *decentralized*, implying complete autonomy of the subsystems. However, some central control is often needed to avoid disorder [132]. The system may also be *distributed*. In this case, the subsystems share sensing information with at least their nearest neighbors [133] using either a shared global memory or message passing through a control channel [134].

Communications between the levels should be restricted to the essential. This general principle is known as *subsidiarity* or *loose* or *weak coupling*, which is the best and most efficient way to organize hierarchy [47], [95], [96]. In subsidiarity, centralized control is used, but the subsystems are relatively autonomous. Decisions are made near the problems. A related concept is *orthogonality*, which means isolating subsystems to minimize interference between them [113].

Many systems having nonlinear relationships between the parts are mathematically intractable. In general, a system is mathematically tractable and can be analyzed when the coupling or interaction between the subsystems is nonexistent, weak enough to be ignored, or linear. In these cases, the superposition theorem is valid [18]. A special form of nonlinear systems is affine systems that are otherwise linear but they have an additive term at the output [135]. Affine and other simple nonlinear systems are tractable. However, even affine systems require nonlinear solutions [113], not to mention more complex nonlinear problems. If the nonlinear coupling is strong enough, emergent phenomena may appear. Nonlinear relationships can hence lead to chaotic behavior if not properly designed. Subsidiarity and orthogonality decrease the number of relationships between the parts of a system and thus facilitate tackling this challenge.

### 4) HIERARCHY OF SYSTEMS

We can classify systems into the general hierarchy as shown in Figure 5, that we formed by combining the results in [41], [48], [63]. A more comprehensive hierarchy is presented in [63], combining hierarchy and evolution. Similar general hierarchies are included in [47], [136], [137]. A hierarchical and distributed version of the system model (Figure 4) is valid for all these systems. Still, in some cases, open-loop control without sensors may be needed to support fast enough although rough responses [52].

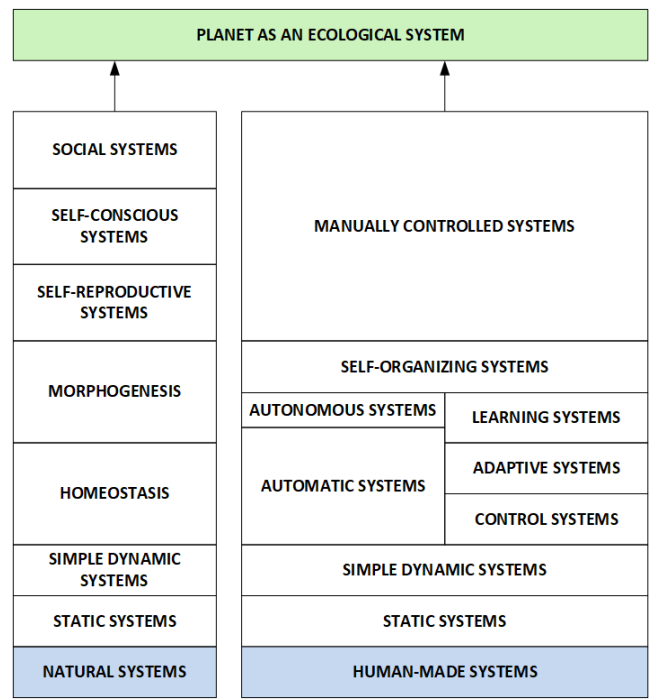

**FIGURE 5.** General hierarchy of systems. The uppermost ecological system includes natural and human-made systems.

A simplified hierarchy includes automatic, autonomous, self-organizing, and manually-controlled (often remote-controlled) systems, from bottom up. In large systems, several of these systems might be combined to manage complexity. *Automatic systems* do not need any manual intervention but may receive some external control information such as a set-point value, reference signal, or route to a given goal. Automatic systems include control systems, adaptive systems, and some learning systems that use external control signals. *Autonomous systems* are learning systems that do not need any external control during operation except for the initially given goal. *Self-organizing systems* are autonomous systems able to change their structure.

In *manually-controlled systems*, the decision block is replaced by a human being. The human setting the goals is in the loop, and therefore these are the most intelligent of all systems and belong to self-conscious systems. Generally, some conative structure or freedom of choice is needed to provide goals for a machine [48].

In natural systems, homeostasis corresponds to control and adaptive systems. Morphogenesis, in turn, corresponds to learning and self-organizing systems. Only natural systems are *self-reproductive*, *self-conscious*, and *social*. Natural systems differ in many respects from human-made systems, from robots and other artifacts. For example, natural systems do not use any set-point values or reference signals for control, adaptation, learning, or self-organization. Self-consciousness is a product of evolution, made possible through self-reproduction.

### C. SYSTEM MODELS
*System model*, *system architecture*, and *system platform* [105] concretize the vision; together with the scenarios and system ideas, they give concrete goals for the research. Modern system models in different disciplines include:

- *networked control systems* (NCS) [138], [139] and *autonomous, cooperative, and self-organizing robots* [140]–[142] in control theory [52], [143],
- *distributed and autonomic computing* [128], [144], *mobile agents* and *multiagent systems* (MAS) [145], [146] in computer science [147], [148], and
- *software-defined networks (SDN)* [149], *self-organizing networks (SON)* [150] and *cloud and edge servers* in communication theory [151], [152].

These disciplines have similar trends towards hierarchical distributed systems, and finally, to autonomic or cooperating, autonomous, and self-organizing systems, but use different terminology. Similar situations can be observed between other disciplines as well [18]. Hence, there is a clear need for interdisciplinary discussions, surveys, and vocabularies so that researchers can locate other fields' results, use common, earlier coined terms, and avoid work already done.

A system architecture defines how intelligent systems are divided into parts, and a platform is an execution environment for the software of these systems [105].

## V. FROM VISION TO RESEARCH AND EDUCATION
### A. PLANNING RESEARCH AND EDUCATION
Scientific inquiry has been successfully carried out in a compartmentalized manner in specialized disciplines [125], by selecting at every step the most beneficial research topics, which are so called hot topics. Focus on citations in research evaluation has resulted in stagnation and incremental research. Exploratory research leading to breakthroughs has been widely ignored [153], and therefore, new ideas are harder to find [154]. Furthermore, the reactive greedy approach does not always return optimal or even good solutions, especially when the problem is complex with various nonlinear interactions [50].

Rather than chasing citations, researchers would need persistence, a focused research program, good methodology, and publishing in relevant journals [155]. Funding organizations should encourage such steps by considering innovative new multivariate assessments of research productivity, including assessing social impact. Furthermore, researchers could act proactively and contribute to selecting research topics as was done, for example, by Weiser [10].

We suggest five steps for planning research and education to support realizing the Smart and Sustainable World vision. Table 2 illustrates these steps. We focus on research. Compared with the general system life cycle in systems engineering [22], our suggestion can be located at the concept development stage, though activities similar to the engineering development stage are performed during the actual research when building prototypes.

Planning proceeds from top down, but feedback from bottom up is also essential. The key concepts related to these steps were described in the previous sections. Research environment, enabling technologies, research problems, and education are discussed in more detail below. The planning starts by refining the vision, if necessary, and continues by writing scenarios on how intelligent systems provide services for their users. The scenarios are hence closely related to user needs and goals, as well as human acceptance and control.

When scenarios are written, the most important vertical industries can be considered (Section V-B), as well as the local industry's needs and the core competences of the university [156]. Further information sources include European Commission's missions for Horizon Europe [103], the twelve megatrends [157], [158], ten megatrends for the 2020s [159], Gartner hype cycle [160], and other technology trends, for example [161] and [162]. Foresight tools provide yet one more possibility to consider research that can be expected to have a significant impact [163]. In addition to writing scenarios, this information can be used to estimate the technology that will mature soon and recognize open problems as well.

Scenario writing should be guided by the Sustainable Development Goals and their Key Performance Indicators (KPIs) [164], as sustainability is central in the vision. The White Paper on 6G and UN SDGs [165] proposes how the SDGs can be supported with research on 6G systems. This information is directly applicable, as the future intelligent systems will be 6G systems. The resulting scenarios can be used to draft the impact, whether scientific, societal, economic, cultural, or environmental. Further, they are tools for discovering goals and goal hierarchies for intelligent systems.

The next step, specifying system requirements, is based on the scenarios. Here, performance parameters, basic resources, and the fundamental limits of nature play an important role. The performance requirements can be met with a system specification by applying system ideas and selecting suitable system models. Decisions on organizing the intelligent systems into hierarchical levels and relatively autonomous subsystems are made, for example. These decisions lead to identifying the required functionality in more detail, architecture considerations, and specifying the research environment and enabling technologies (Section V-C).

**TABLE 2.** Steps to plan research and education.

| 1. Smart and Sustainable World vision | Prosperity for the people and the planet is achieved with intelligent systems that sense their environment, make proactive decisions on actions advancing their goals, and perform the actions on the environment. Sustainable development is emphasized in decision making, and system performance is optimized to save basic resources. Humans observe the autonomous operation through user interfaces and, when needed, revise the operation or control the systems manually. |
|---|---|
| 2. Scenarios and use cases | Operation and interaction of intelligent systems with their environment to achieve the selected goals. |
| 3. System requirements | *Functionality*; *stability and scalability*; *performance and constraints*: efficiency of the use of basic resources: materials, energy, information, time, frequency, and space, fundamental limits; *dependability and security*: availability, reliability, safety, confidentiality, integrity, maintainability; *cost*. |
| 4. System specification | *System ideas*: optimization, decision-making, open- and closed-loop control, hierarchy, degree of centralization. *System models*: Networked control system (NCS), automatic, autonomous and self-organizing systems, multi-agent systems, autonomic computing, software-defined networks. *Enabling technologies*: artificial intelligence (AI), computational intelligence (CI), Internet of Things (IoT), cyber-physical system (CPS), high-performance computing (HPC), big data and mobile cloud computing, cybersecurity, smart devices, ultra-densification, energy harvesting, software-defined network (SDN), network function virtualization (NFV), device-to-device (D2D) communications, radio access technology (RAT), massive multiple-input multiple-output (MIMO), millimeter wave (mmW) and THz communications, wireless charging, optical and satellite communications, green communications. *Research environment*: resources and services to conduct research. |
| 5. Research problems and education content | *General research problems*: Massive scaling, architecture and dependencies, creating knowledge, big data, robustness, openness, security, privacy, humans in the loop, emergent behavior. *Knowledge and skills*: Autonomous systems, decision theory, network and traffic theory, nonlinear system theory, systems thinking. |

The last step is to specify the research problems and the education topics. The research problems determine the focus of the research. Requirements lead to research problems when the required functionality or performance cannot be achieved without new solutions. The technologies to be developed, e.g., 5G, AI, or IoT, depend on the selected focus. The open problems listed in Section V-D can be considered as well. However, a random component and scientific curiosity should always be allowed to support the evolutionary approach and encourage creativity in research. The education topics provide the required general knowledge to study system-level problems and apply systems thinking (Section V-E).

These steps produce a plan for contributing to the vision by generating new knowledge and educating experts. The plan contains the ingredients needed to start the actual research as described in Section III. The conceptual analysis is performed during the first four steps. The system specification represents a hypothesis about the system. The general research problem is reduced to subproblems during the fifth step. Moreover, the research environment provides resources and services for experiments, and education content refines education.

We emphasize responsible conduct of research following the values of our society. We highlight justice leading to trust; trust leading to cooperation; cooperation leading to social contracts that form the basis of our society and its safety; and integrity preventing plagiarism and fabricating results. The values are protected by norms and by respecting colleagues. Additional values to follow include motivation (to incite people to act), creativity (to produce something new), criticism (to obtain high quality), freedom (the power to choose what one wants to study), objectivity (correspondence with reality, no bias), openness (to prevent intrigue), communication (clarity of expression), and diversity (to bring in different ideas and perspectives). For more details on research ethics, see [166]. Knowledge should be shared with manuscripts that are critically evaluated before they are published. Inter- and transdisciplinary work should use collective decision making. These values need to be present in education as well; they should be taught to the students and followed by the teachers.

### B. VERTICAL INDUSTRIES
The most important vertical industries in Europe include *factories of the future*, *automotive and mobility*, *healthcare*, *energy*, and *media and entertainment* [167]. Services and products are developed mainly for the vertical industries. As these verticals are also globally significant and can contribute to sustainable development, we focus on them. However, we add *agriculture* to this list as it is responsible for over 11% of global greenhouse gas emissions [168].

A *vertical market* is a group of companies that serve each other's specialized needs and do not serve a broader market [169]. A vertical market is tightly focused on meeting the needs of one specific industry (i.e., application area). In the systems view, a *vertical market includes all abstraction levels* from physical implementation to services, although the markets are named according to the application areas. In contrast, a traditional *horizontal market* sells its goods and services in more than one industry and, therefore, targets a range of business segments.

The fourth industrial revolution introduces IoT and CPS to realize efficient and flexible factories of the future [161].

5G and 6G systems provide a reliable and fast communication channel for connecting all the devices into a single system, combining fiber and wireless, including satellite connectivity. Moreover, 6G introduces AI at the edge, near the data sources and controlled processes. Process optimization inside factories will be crucial [170]. Such optimization is an obvious application for intelligent systems.

The key trends in the automotive industry are connecting vehicles to the Internet and each other and advancing towards higher automation levels. Vehicles collect data with their sensors, change information with other vehicles, and use the road infrastructure as well [171]. Intelligent systems will play a central role in increasing automation and also in realizing cooperative vehicles.

Electronic health (e-health) and mobile health (m-health) offer ICT to healthcare, to transfer health resources and healthcare by electronic means, and to support the medical and public health practice by wireless devices like mobile phones and patient monitoring devices [172]. Although these developments will change healthcare significantly, larger changes will be triggered by introducing IoT, more systematic data collection and sensing, and intelligent systems using the collected data and AI to control the various healthcare related processes.

Reliability is emphasized in the energy sector, as device installations are expected to last 20 years, and the controlled processes are critical for society and the other vertical industries. Grid protection and control and smart metering are examples of the ongoing trends. Distributed generation and storage of power and micro-grids will change the energy sector considerably in the long run [173]. Like other verticals, intelligent and autonomous systems enable low latencies in decision making, managing complex processes, and optimizing resource usage.

The key trend in media and entertainment is the change of user habits and expectations. Media is consumed more on-demand and on-the-move [174]. A major need for intelligent systems will be optimizing network usage, for example, to minimize overall bandwidth usage and server loads for a large number of mobile users by caching media on edge servers.

In agriculture, smart farming introduces new ICT solutions to support efficient production processes [175]. The relevant technologies include satellite imagery, agricultural robots, sensor networks, and unmanned aerial vehicles. Data collection and analysis provide sensor data for intelligent systems controlling the processes.

## C. RESEARCH ENVIRONMENT AND ENABLING TECHNOLOGIES

*Research environments*, also called research infrastructures, are needed for experiments (Figure 1) in testing hypotheses and prototypes. Experiments bring out complex systems' interactions, emergent properties, and end-to-end performance. Research infrastructures provide resources and services for research communities to conduct research and foster innovation [176]. They may be single-sited, distributed, or virtual. They include *"major scientific equipment or sets of instruments; collections, archives or scientific data; computing systems and communication networks; and any other research and innovation infrastructure of a unique nature which is open to external users"* [176]. For research on Smart and Sustainable World, the research environment should contain a platform for intelligent systems and resources for simulations.

*Enabling technologies* are used in the research environments and prototypes to provide the functionality for which mature technologies are available. Research is a combination of theory, simulations, technologies, research environments, experiments, and big data [177]. Inter- and transdisciplinary mission-oriented research towards Smart and Sustainable World needs to be accompanied with a large selection of technologies, as listed in Table 2. This list includes the ten key enabling technologies for 5G systems [178] that will play an important role in realizing the vision. We add artificial intelligence, cybersecurity, and high-performance computing (HPC, or super-computing) among the enablers [179]. We also list cyber-physical systems (CPSs), that is, integrations of computation and physical processes, combining control, computing, and communications [139], as an enabling technology as CPS is a rather generic system model.

Furthermore, data representations, identifiers, naming rules, protocols, and interfaces between subsystems or modules are crucial for the targeted large-scale systems. Technologies of the World Wide Web Consortium (W3C) and Semantic Web support the efficient use of data in such systems and codifying semantics [180]. Many future systems will be widely accessible. This can lead to privacy, security, and safety problems due to possible internal faults and physical or cyber attacks [99]. Sufficient safety and security are crucial and establishing trust among users as well. Private, isolated Internet may sometimes be needed to guarantee cybersecurity.

Computing and control may be distributed. In communication networks, the whole signal path from sensors to the decision block and the actuators must fulfill the reliability and real-time requirements. To avoid delays, processing and communications are often close to the user, as in multi-access edge computing (MEC) servers and device-to-device (D2D) communications. Many implementations will be based on software and virtualization.

Internet of Things (IoT) is an essential enabling technology to connect sensors, actuators, user interfaces, and other components to the Internet. Current IoT platforms offer various services for data storage, data access control, data analysis, resource discovery, etc. The number of available platforms is large. For example, the authors in [181] survey 39 platforms and in [182] 40 leading IoT solutions. IoT will have a significant role in all vertical industries, as IoT systems will be pervasive - the number of IoT devices is expected to exceed 75 billion worldwide by 2025 [183]. The market size of IoT has been estimated to grow, from 100 billion U.S. dollars

**TABLE 3. Examples of open problems.**

| Field | Open problems |
|-------|---------------|
| General | Privacy protection in the Internet of Things. Wearable device development and security challenges. Bridging the gap between the world and its computer model. How to process heterogeneous big data from smart things. Ubiquitous computing and security. Smart software infrastructure for smart worlds. How to evaluate the intelligence of the smart world [188]. |
| Collaborative systems | Autonomy, sensing of environment and neighbors, cooperation, and precise communication [189]. |
| Artificial intelligence | Optimizing AI's economic impact, law and ethics, robust AI [190]. Frame problem [191], [48], [65], [192]. |
| Security | Risks of cyber-attacks, defensive schemes, and integrated evaluation platforms to protect IoT-based systems against cyber-attacks [193]. |
| Human control | Transparent and controllable intelligent systems. General theory for meaningful human control over autonomous systems [78]. |
| Internet of Things | Massive scaling, architecture and dependencies, creating knowledge and big data, robustness, openness, security, privacy, and humans in the loop [9]. Interoperability, scalability, flexibility, energy efficiency, mobility management, and security [194]. |
| Networked control systems | Time delays, packet losses and disorder, time varying transmission, competition of multiple nodes, data quantization, clock asynchronism, and network security and safety [195]. Networking technologies, fault-tolerant control, bandwidth allocation and scheduling, and integration of components [138]. Coding for robustly stable control, time-varying data rate, nonlinear feedback, routing [196]. |
| Green IoT | Energy efficient system architecture, energy efficient service composition strategies, situation and context awareness regarding users and applications, energy efficient WSN management, energy efficient cloud management [197]. |

market revenue in 2017 to 212 billion by the end of 2019 and around 1.6 trillion by 2025 [184].

Miniaturization leads to the Internet of Nano-Things, where the power consumption is reduced by orders of magnitude to microwatts, nanowatts, and picowatts [185], resulting in smaller devices and new applications. Internet of Nano-Things and miniaturization support developing devices that are embedded in the environment and harvest energy for their operation [186]. However, the fundamental limitations of electronics challenge this trend. Smaller power also implies lower computation rate and simpler processing. 5G and 6G systems will also be important enablers and research topics. They bring very short latencies in the order of 1 ms and reliable communication, enabling wider use of wireless communications in all vertical industries, for example, to connect the devices forming the feedback control loop.

### D. OPEN PROBLEMS

The open problems presented in this section can be considered as concrete goals that guide the research [1], [2], [72]. The open problems listed in the literature often have a broad scope and need to be narrowed. The intractability of multiple causal feedback loops is an example of a general problem related to the Smart and Sustainable World vision, leading to the need for higher-order cybernetics [18], [187]. The three-body problem in physics is a famous example on the intractability of the many connected feedback loops, where the nonlinear interactions between the bodies generate emergent behavior [18].

Table 3 presents examples of open problems. In artificial intelligence, the frame problem is one of the most difficult open problems. This problem is about the inability of machines to understand semantics, that is, relationships between abstract symbols and reality [191]. No theory exists for this problem and therefore no machine understands semantics [48], [65], [192]. Understanding semantics is an important part of self-consciousness as well as cognition. The term cognitive has been overused in recent years to refer to learning, as in cognitive radios. Russel *et al.* [190] list open problems requiring solutions to guarantee artificial intelligence that is beneficial for society and robust (i.e., behaves as intended). Robust AI includes meaningful human control discussed in Section II-B and is closely related to technology acceptance as well.

The open problems for IoT systems [9] overlap partially with the technology enablers for 5G systems. Research problems related to IoT, specifically to the intelligent use of IoT resources, will be crucial due to the expected large number of IoT devices in the verticals. Additional open problems in communications can be found from [198], [199]. Open problems in localization are discussed in [200].

### E. EDUCATION

Research on intelligent systems realizing the Smart and Sustainable World vision requires wide expertise. We emphasize systems thinking and systems engineering and concentrate on natural sciences, technology, engineering, and mathematics (STEM) topics. We suggest the content listed in Table 4 to prepare students for the systems approach and interdisciplinary projects. This content was inspired partially by [111]. Our focus is on updating the curricula of engineering degree programs that already provide the required preliminary knowledge and skills for studying the suggested content. State of the art technologies such as AI and IoT are covered in the education as planned in each degree program. Here, we focus on system-theoretical topics.

In addition to the broad set of technological subjects, developing technology for humans and considering the societal dimension call for social sciences and humanities (SSH). Additional expertise needed in inter- and transdisciplinary research can be built in research teams after graduation. Working in such teams can be practiced during studies in interdisciplinary project courses. Education curricula are needed for continuing (i.e., life-long) learning as well, but this is outside the scope of this article. Many computing curriculum efforts already include systems thinking and systems engineering [201]. A comprehensive guideline is included in [202], and the Systems Engineering Body of Knowledge (SEBoK) [38] contains a wealth of information.

**TABLE 4.** Suggested system-theoretical subjects.

| Subject | Contents | References |
|---------|----------|-----------|
| Autonomous systems | Positive and negative feedback, stability, hierarchy, degree of centralization, automatic control, adaptive systems, autonomous learning, self-organizing systems, federated learning, autonomous agents and robots, cooperation and competition, networked control systems, history | Albus 2001 [46], Bekey 2005 [141], Bernstein 2002 [143]. Coulouris 2012 [129], Dorf 2017 [204], Ghosh 2015 [205], Haykin 2013 [206], Kotseruba 2020 [207], Maurer 2015 [208], Mesarovic 1970 [125], Murata 2012 [142], Ogata 1995 [209], Ogata 2010 [52], Russell 2010 [49], Sheridan 2016 [210], Tanenbaum 2007 [128], Valavanis 2007 [211], Widrow 1985 [212], Yang 2019 [213] |
| Optimization and decision theory | Traditional optimization, multiobjective optimization and multiple-criteria decision making, conventional artificial intelligence, computational intelligence, deductive logic, pattern recognition, statistical methods, history | Chaturvedi 2008 [214], Coello 2007 [215], Bishop 2006 [216], Eberhart 2007 [217], Figueira 2005 [218], Gass 2005 [219], Köksalan 2011 [220], Konar 2005 [221], Marsland 2015 [57], Medsker 1995 [59], Michalewicz 2004 [50], Mitchell 2009 [222], Talbi 2009 [223], Tsoukias 2008 [224], Yang 2010 [225] |
| Network and traffic theory | Information theory, estimation theory, queueing theory, random graphs and stochastic geometry, history | Haenggi 2013 [226], Newman 2018 [227], Shortle 2018 [228] |
| Nonlinear system theory | Nonlinear models, Volterra series, memory polynomials, nonlinear compensation, nonlinear control, nonlinear dynamics, solitons and chaos, Lyapunov stability, history | Haddad 2008 [229], Perona 2005 [230], Schetzen 2006 [231], Strogatz 2014 [232] |
| Systems thinking | System principles, sustainability and resilience, general hierarchy of systems, system dynamics, complexity theory and self-organization, evolutionary dynamics, stability, scalability, efficiency, deductive logic and pattern recognition, statistical methods, emergence, system archetypes, tragedy of the commons, subsidiarity and loosely coupled systems, system traps, places to intervene in a system, computational complexity, measurement theory, reliability engineering, dependability and security, fundamental limits, methods and tools for systems thinking, history | Arora 2009 [233], Avizienis 2004 [99], Barrow 1998 [100], Benbya 2006 [95], Bertalanffy 1971 [18], Bossel 2007 [47], Boulding 1985 [63], Boulton 2015 [234], Checkland 1999 [20], Cockshott 2015 [235], Dewdney 2004 [117], Elsayed 2021 [236], Epstein 2019 [93], Erdi 2008 [237], Forrester 2007 [238], [239], Francois 1999 [187], Hubka 1988 [64], Kahlen 2017 [240], Klein 1990 [24], Kossiakoff 2020 [22], Liu 2016 [241], Markov 2014 [118], Meadows 2008 [2], Mesarovic 1970 [125], Nielsen 1993 [73], Nielsen 2015 [17], Norman 2013 [242], Nowak 2006 [67], Ramage 2020 [85], Repko 2020 [102], Richardson 1991 [98], Senge 2006 [243], Sterman 2000 [124], Trivedi 2017 [116], Tzafestas 2018 [112], Yanofsky 2013 [244] |

An effective curriculum should not be made up of independent courses that leave the difficult task of their synthesis to the student [85]. The curriculum should offer enough general knowledge to students and provide an integrated hierarchical worldview where development is evolutionary and based on generalized feedback from the environment as in [63]. This can require a hard-to-be-achieved paradigm shift from a purely reductive approach using multidisciplinarity to a combination of reductive and systems approaches using inter- and transdisciplinarity. As suggested in [2], [63], [98], [124], the concepts of hierarchy and feedback are essential tools in most sciences, including social sciences. Thus, education should familiarize all students with these concepts. Teaching a worldview is more important than teaching disconnected fast changing facts. A worldview integrates facts into a whole and usually changes slowly.

The compulsory courses should include enough material on mathematics, physics, signal and circuit theory, electronics, control theory, computer science, software engineering, and communication theory. The target is to avoid isolating students in separate compartments. Instead, with overlapping knowledge in the curricula, future interdisciplinary projects become manageable. Research methods must have a sufficient role in the curricula as they are an essential part of our research culture. Sustainable Development Goals need to be considered as well; to educate experts that can contribute to sustainable development [203]. Social and human sciences need to be introduced in sufficient detail to support broad inter- and transdisciplinary work and contributing to sustainable development. The Smart and Sustainable World Vision and the research related to the vision (as described in this article) can be presented to illustrate the type of contributions the students can make on their careers and justify their studies' content.

The reasons for and dependencies of courses should be explained, for example, using diagrams. Specialists have the expertise to teach the courses. The bottom-up reductive approach should be combined with the top-down systems approach, in this order, by giving links to history and system ideas. That is, the combination of reductive and systems approaches used in research should be used in education as well. History and analogies can make the topic more interesting to students appreciating systems thinking [20], [93], [98]. Analogies are efficient in learning new subjects and developing conceptual thinking to gain transferable knowledge to manage the world. A preliminary systems view in the beginning may be useful. In addition to analogies, complexity can be managed using hierarchies and abstractions that ignore nonessential details through idealizations.

Students may find separate system-theoretical courses demanding; hence, the material could be included in the existing and carefully selected new technical courses supporting interdisciplinary systems thinking. A separate system-theoretical course using horizontal integrative learning is most useful during doctoral studies or at the end of master's studies after appropriate vertical integration using systems view has been made in each course. One must know the parts that are going to be integrated, and therefore horizontal integration transcending the borders of courses should not be started too early.

Examples of horizontal integrative learning include phenomenon-based, problem-based, project, situated, and authentic learning, and thesis work. They all target meeting societal and educational demands and focus on authentic and fundamental research problems [245]. Such research problems are expected to lead to tangible outcomes. The problems should be presented thoughtfully and engaged collaboratively to generate multiple possible solutions. A deep approach to learning should be used. The aim is that the learners can discover that knowledge is tentative and open to continued shifts and changes.

## VI. RECOMMENDED ACTIONS

We recommend actions to start refining ICT research and education so that universities can contribute to realizing the Smart and Sustainable World vision by generating new knowledge and educating experts. We propose collecting experts to participate in the effort, networking, collecting background material, focusing the vision, reviewing the research profile and environment, and reviewing the education curricula. This set of actions was inspired by [111], [246].

The research team performing the actions should have a wide range of expertise. In addition to STEM, social sciences and humanities (SSH) are needed. Broadly, to correctly consider the human and social dimensions, one needs to analyze and understand *"different aspects of politics, social and cultural norms, ethics and legal frameworks, production and consumption patterns, traditions and lifestyles, and historic trajectories"* [247]. As an example, the required wide range of expertise is recognized in the EU's Ethics Guidelines for Trustworthy AI [28] stating that trustworthy AI *"requires a holistic and systemic approach, encompassing the trustworthiness of all actors and processes that are part of the system's socio-technical context throughout its entire life cycle."*

The recommended actions are summarized in Table 5. The first recommended action for a university is to form a working group to support both research and education. The group participates in the rest of the actions by making plans, sharing experiences, providing lectures, and collecting material. Experiences are shared to give insights into the potential gains to commit the key personnel to the actions. To avoid extra administration and meetings, the working group could be aligned with the existing management. Since the members will be mostly professors and senior staff members, they will naturally be responsible for implementing the actions.

**TABLE 5.** Recommended actions.

| |
|---|
| 1. Form a working group |
| 2. Use tutors as advisors of young researchers |
| 3. Participate in networks |
| 4. Collect background material |
| 5. Plan the university's role in realizing the vision |
| 6. Review the university's research profile |
| 7. Develop the research environment |
| 8. Review the university's education curricula |

The second recommended action is to nominate tutors for young researchers. Basically, all doctors and doctoral students should act as tutors to younger students; this should be a mandatory part of their education. The tutors are educated to become next generation system thinkers so that this idea will continue. The third action is to participate in selected networks to collaborate, obtain the-state-of-the-art knowledge, and gain visibility. Through collaboration, the university can focus on its strengths and agree on the division of responsibilities within the network in both research and education. Networks such as public-private partnerships facilitate building critical mass to be attractive partners and generate significant results. Networks are preferred by funding organizations as well.

The fourth recommended action is to collect background material in the form of annotated bibliographies, chronologies, and vocabularies. Background material for research and education supports using the existing body of knowledge and avoiding work already done. Vocabularies form the basis for conceptual analysis and include definitions of terms, synonyms, antonyms, taxonomies, hierarchical relationships to other terms, and acronyms as in a thesaurus. Translations into students' native language are also useful for learning. The starting point could be [105]. In addition, the IEEE has a thesaurus and a hierarchical taxonomy based on the thesaurus to unify terminology [248], [249].

The fifth action is to plan the university's role in realizing the Smart and Sustainable World vision, the first four steps from Table 2. This plan covers the scenarios and the initial system specification, including definitions of enabling technologies and the research environment. The plan should look proactively about ten years to the future [46]. The design should be kept at a sufficiently high level; more details can be added during the following actions and later during the actual research and education as needed.

The sixth action is to review the university's current research profile and initiate the required new openings. This is the research part of the last step in Table 2. Planning the research to solve the research problems produces a coherent project portfolio to renew the research and contribute to the vision. This portfolio can form the basis for a research platform. The required expertise indicates the disciplines needed in interdisciplinary research. In the STEM domain, knowledge is needed from mathematics, physics, chemistry, electronics, control theory, computer science, communication theory, and general system theory when complete systems are developed, ICT performance is improved (e.g., with faster electronics), ICT is integrated into physical objects, and new materials for ICT components are developed, for example. Similarly, social sciences and humanities bring their expertise as required, for example, to guarantee that usefulness, usability, acceptance, and meaningful human control are considered in sufficient detail, as discussed in Section II. The identified new expertise calls for new openings and recruitments. For the actual research, we stress combining the reductive and systems approaches, as described in Section III.

The seventh action is to develop the research environment (Section V-C) to conduct research and perform experiments. The research environment can be used in both education and research to test autonomous and remote-controlled systems and their components. Students can be familiarized with the technology and perform realistic experiments, and researchers can obtain more valuable results. In addition to testing individual intelligent systems, testing their co-existence on a shared platform can provide valuable insight.

Moreover, deploying research results into a realistic environment supports collaboration with companies and commercialization, thus amplifying the impact of the research. The three conventional research methods are analysis, simulations, and experimental research. Research is typically more abstract and general at universities, whereas industry focuses on more concrete research and development. Thus, in principle, the approaches can complement each other, but only when a common language can be found. This is possible using abstraction as explained in [93]: ''The ability to move freely, to shift from one category to another, is one of the chief characteristics of 'abstract thinking.''' A similar problem can be observed between different scientific disciplines because reductive tradition results in fragmentation of science and terminology. These problems can be solved by bringing systems thinking into university curricula. Systems thinking operates in the space between universities and industry and assumes the attitudes of both [111]. System ideas such as the feedback concept should be included in social scientists' education; see [98]. Engineers are designing their systems for society, and thus they should, in turn, be aware of the best results of social sciences.

We suggest that researchers define for the intelligent systems common structures that produce characteristic behaviors, that is, *system archetypes* that have survived the test of time [2]. These archetypes can be collected, together with their simulation models, into a *system zoo* [47] that can be provided to researchers, teachers and students as a part of the research environment, as for example in [250]. The system zoo can be used in courses and research projects and to introduce this knowledge to new disciplines as well.

The eighth and last recommended action is to review the university's ICT education curricula and build the required new content. This is the education part of the last step in Table 2. Weighting between the compulsory topics listed in Section V-E depends on the degree program. Horizontal integrative learning can be supported with project courses that collaborate with the research projects, thus preparing students for research careers. The research projects provide a good source of examples for all courses in the curricula. Moreover, presenting the research projects' role in realizing the Smart and Sustainable World Vision gives concrete examples on the type of contributions researchers can make on their careers.

As discussed above, the aim is to define common goals and build a coherent project portfolio and education content. Common goals result when the research team performs planning together. The research portfolio can be built based on this work. In education, the refined curricula prepare the students for the systems approach and interdisciplinary projects. Research and education produce feedback to refine the plan.

## VII. CONCLUSION

We are convinced that common goals, including shared visions and research problems, promote teamwork and result in coherent project portfolios, thus improving the research culture. We suggest the Smart and Sustainable World vision, steps to plan research and education based on this vision, and actions for realizing the plan. Universities are the main actors in this plan, generating new knowledge and educating experts. Generally, Higher Education Institutes (HEIs) should contribute to SDGs by producing and sharing relevant knowledge [203].

Intelligent systems are at the core of the vision. These systems fulfill the needs of society, individuals, and industries but also optimize resource usage as intelligent use of the limited basic resources will be crucial to realize sustainability. Scenarios are used to define system requirements and finally to specify the systems. Human acceptance and control are emphasized. Systems must have high efficiency, availability, reliability, safety, and security. Research and education produce feedback to support evolutionary development and encourage creativity in research.

The principle of subsidiarity is an efficient way to organize hierarchy to minimize the nonlinear relationships between the parts. The closed-loop feedback concept is essential to produce intelligent behavior, that is, to advance given goals in an uncertain environment. As setting goals implies self-consciousness, humans are needed to set goals for intelligent systems when the systems are developed and during operation. We also suggest that humans stay in control and revise the operation of the intelligent systems when necessary.

Inter- and transdisciplinary effort is required to cover the wide expertise and synthesize the research results into a coherent whole. Research on intelligent systems requires expertise from control theory, computer science, and communication theory. Social and human sciences and several other disciplines are needed as well, as the intelligent systems coexist with humans and serve them.

We emphasize the systems view complementing the conventional reductive view as a new research paradigm. The systems view is not a replacement for the reductive view. The reductive view proceeds from simple to complex. The emphasis is on details at the expense of the whole. The system is assumed to be isolated from the environment. This is a good approach to start with, both in education and research. However, when systems are complex, or there is a strong nonlinear coupling between the parts and the environment, the problem is easily intractable, and systems view is called for. Moreover, local optimization does not lead to global optimization, which is attempted in systems thinking.

The need for systems view has been widely recognized, and it is necessary both in the mission-oriented approach and to generate a large impact. Systems thinking is a form of generalized inference that is needed to replace deduction in mathematically intractable problems. Understanding history is an essential part of the systems view and our general knowledge. The quality of research is improved by a broad education, for which we have recommended concrete actions to prepare the students for the systems approach and interdisciplinary projects. There is a clear need for interdisciplinary surveys and vocabularies so that researchers can locate other fields' results, use common, earlier coined terms, and avoid work already done.

Research and education should proceed from bottom-up reductive experimental-inductive to top-down hypothetico-deductive and systems approach, in this order, since this is the most natural way of learning. A preliminary systems view in the beginning is useful. This view can be created starting from our Smart and Sustainable World vision.

A concrete next step in realizing this vision and strengthening the role of the systems approach in research and education would be to start performing the recommended actions and refine the suggested approach as the work proceeds and experience is gained. The future areas of research will be defined during this process, as well as the development of education curricula. The focus of the research is determined by specifying the research problems. System requirements lead to research problems when they cannot be met with current solutions. The open problems presented in the literature can also be considered though they often have a broad scope and thus need to be narrowed. We emphasize system-theoretical topics in education. In addition, state-of-the-art technologies are covered as planned in each degree program. The final goal of both research and education is to reach the Smart and Sustainable World vision, whose fundamental goal is prosperity for people and the planet, now and into the future.

**ACKNOWLEDGMENT**

Discussions with Antti Anttonen, Adrian Kotelba, Kari Leppälä, and Mahnaz Sinaie are gratefully acknowledged.

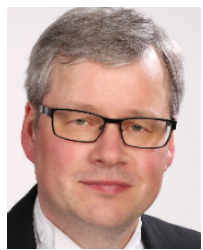

**JUKKA RIEKKI** (Member, IEEE) received the D.Sc. (Tech.) degree from the University of Oulu, Finland, in 1999. He visited the Electrotechnical Laboratory, Intelligent Systems Division, Tsukuba, Japan, from 1994 to 1996. Since 2001, he has been a Professor of Software Architectures for Embedded Systems with the University of Oulu. Since 2014, he has also been the Dean of the Faculty of Information Technology and Electrical Engineering. In his Ph.D. thesis, he studied reactive task execution of mobile robots. In 2000, he started to study context-aware and ubiquitous systems. He has been an advisor of ten doctoral theses. He has published 45 journal articles, six book chapters, and over 180 conference papers. His research interests include the Internet of Things and edge computing, emphasizing distributed and resource-limited systems.

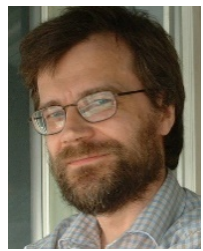

**AARNE MÄMMELÄ** (Senior Member, IEEE) received the D.Sc. (Tech.) degree (Hons.) from the University of Oulu, in 1996. He was with the University of Oulu from 1982 to 1993. In 1993, he joined VTT Technical Research Centre of Finland Ltd., Oulu. Since 1996, he has been a Research Professor of Wireless Communications. He visited the University of Kaiserslautern, Germany, from 1990 to 1991, and the University of Canterbury, New Zealand, from 1996 to 1997. Since 2004, he has also been a Docent (equivalent to an Adjunct Professor) with the University of Oulu. He has given lectures on research methodology at the University of Oulu for about 20 years, including the systems approach. He has been an advisor of ten doctoral students and published over 40 journal articles, 14 book chapters, and almost 100 conference papers. From 2016 to 2018, he was a member of the Research Council of Natural Sciences and Engineering in the Academy of Finland. From 2014 to 2018, he was an Editor of the IEEE WIRELESS COMMUNICATIONS.

• • •